\documentclass[11pt,a4paper]{article}
\usepackage[hyperref]{emnlp2017}
\usepackage{times}
\usepackage{latexsym}

\usepackage{url}
\usepackage[nomessages]{fp}
\usepackage{multirow}
\usepackage{xspace}

\usepackage{amssymb,amsmath,epsfig}
\usepackage{proof}
\usepackage{amsthm}
\usepackage{mathrsfs}
\usepackage{xspace}
\usepackage{algorithm}
\usepackage{amsmath}
\usepackage{graphics}
\usepackage{graphicx}
\usepackage{subcaption}

\newcommand{\notes}[1]{}%{\it {\small {#1}}}}

\theoremstyle{definition}

\theoremstyle{plain}

\newcommand{\ith}[1]{\ensuremath{i^{{th}}}}

\newcount\permx
\newcount\permy
\def\permdot#1#2{
\permx=#1 \advance\permx by-1
\permy=#2 \advance\permy by-1
\psframe[fillcolor=black, fillstyle=solid]
(\permx,\permy)(#1, #2)
}

\newcommand{\boxnum}[1]{{\setlength{\fboxsep}{1pt}\raisebox{1pt}{\hspace{1pt}\fbox{\tiny #1}\hspace{1pt}}}}
\newcommand{\ind}[1]{\ensuremath{_{\kern-0.5pt\boxnum{#1}}}}

\def\namecite{\newcite}

\newcommand{\smallnt}[1]{\ensuremath{_{\mbox{\tiny PP}}}\xspace}

\newcommand{\pseudocode}{Algorithm}
\floatname{algorithm}{\pseudocode}

\iffalse

\else

\fi

\newcommand{\ter}{{\sc Ter}\xspace}
\newcommand{\bleu}{{\sc Bleu}\xspace}

\def\roundposition{1}
\def\roundpositiond{2}
\def\roundpositiont{3}
\edef\rounded{0}
\newcommand{\rdm}[1]{\edef\rounded{0}\FPeval\rounded{round(#1,\roundposition)}\rounded}
\newcommand{\rdmd}[1]{\edef\rounded{0}\FPeval\rounded{round(#1,\roundpositiond)}\rounded}
\newcommand{\rdmt}[1]{\edef\rounded{0}\FPeval\rounded{round(#1,\roundpositiont)}\rounded}
\newcommand\BLEU{\textsc{Bleu}\xspace}
\newcommand\TER{\textsc{Ter}\xspace}

\emnlpfinalcopy % Uncomment this line for the final submission
\title{Attention-based Vocabulary Selection for NMT Decoding}

\author{Baskaran Sankaran and Markus Freitag and Yaser Al-Onaizan \\
IBM T.~J.~Watson Research Center \\
1101 Kitchawan Rd, Yorktown Heights, NY 10598 \\
{\tt \{bsankara,freitagm,onaizan\}@us.ibm.com} \\}

\date{}

\begin{document}
\maketitle
\begin{abstract}
Neural Machine Translation (NMT) models usually use large target vocabulary sizes to capture most of the words in the target language.
The vocabulary size is a big factor when decoding new sentences as the final softmax layer normalizes over all possible target words.
To address this problem, it is widely common to restrict the target vocabulary with candidate lists based on the source sentence.
Usually, the candidate lists are a combination of external word-to-word aligner, phrase table entries or most frequent words.
In this work, we propose a simple and yet novel approach to learn candidate lists directly from the attention layer during NMT training. 
The candidate lists are highly optimized for the current NMT model and do not need any external computation of the candidate pool.
We show significant decoding speedup compared with using the entire vocabulary, without losing any translation quality for two language pairs.
\end{abstract}

\section{Introduction}
Due to the fact that Neural Machine Translation (NMT) is reaching comparable or even better performance compared to the statistical machine translation (SMT) models \cite{jean+:2015,luong+:2015a}, it has become very popular in the recent years~\cite{kalchbrenner+blunsom:2013,sutskever+:2014,bahdanau+:2014}.
With the recent success of NMT, attention has shifted towards making it more practical. Compared to the traditional phrase-based machine translation engines, NMT decoding tends to
be significantly slower. One of the most expensive parts in NMT is the softmax calculation over the full target vocabulary. Recent work show that we can restrict the softmax to a subset of likely candidates given the source. The candidates are based on a dictionary built from Viterbi word alignments, or by matching phrases in a
phrase-table, or by using the most frequent words in the target language. In this work, we present a novel approach which extracts the candidates during training
based on the attention weights within the network. One advantage is that we do not need to determine the candidates with an external tool and also generate
a reliable candidate pool for NMT systems whose vocabularies are based on subword units. The risk with Viterbi alignments is that we could miss some words in the target that are not
fully explained by a Viterbi word alignment. As we train the candidate list with the model parameters, the candidate list is highly adapted to the current model which makes it very unlikely that we miss a high scoring word due to the candidate restriction. 
In this work, we show that it is sufficient to use only the top 100 candidates per source word and speed up decoding by up to a factor of 7 without losing any translation performance.

\section{Neural Machine Translation}
\label{sec:nmt}
%As shown in Figure~\ref{fig:att},
The attention-based NMT~\cite{bahdanau+:2014} is an encoder-decoder network.  
The encoder employs a bi-directional RNN to encode the source sentence ${\bf{x}}=({x_1, ... , x_l})$ 
into a sequence of hidden states ${\bf{h}}=({h_1, ..., h_l})$, where $l$ is the length of the source sentence.
Each $h_i$ is a concatenation of a left-to-right $\overrightarrow{h_i}$
and a right-to-left $\overleftarrow{h_i}$ RNN:
%\begin{equation}
\[
h_{i} = 
\begin{bmatrix}
\overleftarrow{h}_i \\ 
\overrightarrow{h}_i \\
\end{bmatrix}
=
\begin{bmatrix}
\overleftarrow{f}(x_i, \overleftarrow{h}_{i+1}) \\
\overrightarrow{f}(x_i, \overrightarrow{h}_{i-1}) \\
\end{bmatrix}
\]
%\end{equation}
where $\overleftarrow{f}$ and $\overrightarrow{f}$ 
are two gated recurrent units (GRU) introduced by~\cite{cho+:2014_gru}.

Given the encoded ${\bf h}$, the decoder predicts the target translation
by maximizing the conditional log-probability of the 
correct translation ${\bf y^*} = (y^*_1, ... y^*_m)$, where 
$m$ is the length of the target. At each time $t$, 
the probability of each word $y_t$ from a target vocabulary $V_y$ is:
\begin{equation}
\label{eq:py}
p(y_t|{\bf h}, y^*_{t-1}..y^*_1) = g(s_t, y^*_{t-1}, H_{t}),
\end{equation}
where $g$ is %a nonlinear function of 
a two layer feed-forward network ($o_t$ being an intermediate state) 
over the embedding of the previous target word ($y^*_{t-1}$),  
the decoder hidden state ($s_t$), and the weighted sum of encoder states ${\bf h}$ ($H_{t}$). A single feedforward layer then projects $o_t$ to the target vocabulary and applies softmax to predict the probability distribution over the output vocabulary.

%Before we compute $s_t$ and $H_t$, we first covert $s_{t-1}$ and 
%the embedding of $y^*_{t-1}$ into an intermediate state $s'_t$ with a GRU $u$ as:
We compute $s_t$ with a two layer GRU as:
\begin{equation}
s'_t = u(s_{t-1}, y^*_{t-1}).
\end{equation}
%Then we have $s_t$ as:
\begin{equation}
s_t = q(s'_{t}, H_{t})
\end{equation}
%where $q$ is another GRU. 
\noindent where $s'_t$ is an intermediate state.
The two GRU units $u$ and $q$ together with the attention constitute the conditional GRU layer. And $H_{t}$ is computed as:
\begin{equation}
H_t = 
\begin{bmatrix}
\sum_{i=1}^{l}{(\alpha_{t,i} \cdot \overleftarrow{h}_i)} \\
\sum_{i=1}^{l}{(\alpha_{t,i} \cdot \overrightarrow{h}_i)} \\
\end{bmatrix},
\end{equation}

\footnotetext{Same as the decoder GRU introduced in session-2 of the dl4mt-tutorial: https://github.com/nyu-dl/dl4mt-tutorial/tree/master/session2.}

The attention model (in the right box) is a two layer feed-forward network $r$, with $A_{t,j}$ being an intermediate state and another layer converting it into a real number $e_{t,j}$. The alignment weights $\alpha$, are computed from the two layer feed-forward network $r$ as:
\begin{equation}
\alpha_{t,i} = \frac{\exp\{r(s'_{t}, h_{i})\}}{\sum_{j=1}^{l}{\exp\{r(s'_{t}, h_{j})\}}}
\end{equation}

$\alpha_{t,j}$ are actually the soft alignment probabilities, denoting the probability of aligning the target word at timestep $t$ to source position $j$.

\section{Our Approach}
In this section we describe our approach for learning alignments from the attention.

\subsection{Learning Alignments from Attention}
\label{ssec:learn_aligns}
At each time step $t$ of the decoder, the attention mechanism determines which word to \textit{attend} to based on the previous target word $y_{t-1}$, decoder hidden state $s_{t-1}$ and the source annotations $\overrightarrow{h_i}$ and $\overleftarrow{h_i}$. This attention implicitly captures the \textit{alignment} between the target word to be generated at this time step and source words. We formalize this implicit notion into \textit{soft alignments}, by aligning the generated target word $y_t$ to the source word(s) \textit{attended to} in the current timestep. The strength of the alignment is determined by the weight of the attention weights $\alpha_{tj}$. While, the attention weights are probabilities, we treat them as fractional counts distribution over source words for the current target word. %Intuitively this corresponds to the distribution over different source words for a given target word.

Our method simply accumulates these (normalized) attention weights into a  matrix as the training progresses. A naive implementation of this would need a matrix of dimensions $|V_s| \times |V_t|$, which would be infeasible in the typical memory available. Instead, we maintain a sparse matrix to keep track of these raw word counts, where we only update the cells that are touched by the alignments observed in each minibatch. We further delay the accumulation of alignments during the first epoch of training, to ensure that the network can produce reasonably good alignments. And finally, we also employ a threshold $\alpha_{thr}$ and only record the alignments where the attention weights are larger than this threshold. This filters out large number of spurious alignments especially for frequent words, which are unlikely to be of any use.

It should be noted that, the idea of treating the attention weights as soft alignments is already being used in certain cases during decoding. For example, it is a standard practice to get the alignments for the UNK tokens in the decoder post-processing in order to replace it with appropriate target translation using external alignments such as Model-1~\cite{jean+:2015,luong+:2015a,mi+:2016a,lhostis+:2016}. Some of these works, notably~\namecite{jean+:2015} and~\namecite{mi+:2016a} have relied on the alignments generated by external aligners to identify candidate vocabulary for softmax for each training sentence. Additionally, we also propose a way for using these alignments during training. %However, this is the first time that attempts to learn the alignment matrix during NMT training. Additionally, we use these alignments for vocabulary selection in the later stages to speedup the NMT training.

An attractive aspect of our approach is that the alignments could be learned even for previously trained models by continuing the training for one or two epochs. As we show later, it is usually sufficient to learn alignments by accumulating the attention weights for just one additional epoch (see Section~\ref{ssec:results}).

\subsection{Vocabulary Selection for Decoding}
\label{ssec:decoding_voc_manip}
As mentioned earlier, vocabulary selection to speedup decoding has been widely employed in NMT~\cite[\textit{inter alia}]{jean+:2015,mi+:2016a}. In this work, we use the alignments that are learned during training for vocabulary selection. It should be noted that the accumulated attention weights are fractional counts and not probabilities. Secondly, these counts as learned from the attention characterize \textit{target} to \textit{source} alignments. During decoding we are interested in getting the target vocabulary given the source words. So, we first normalize the distribution matrix along target axis and then use the normalized distribution to obtain top-$n$ target words for each source token.

This obviates any need for external tools for generating alignments and also simplifies the decoding pipeline. Following the findings by~\namecite{lhostis+:2016}, we only rely on learned alignments and do not use top-$k$\footnote{It is typical to set $k$ to be 2000~\cite{jean+:2015,mi+:2016a}.} most frequent words or any other resource for decoding. Our experiments (see Section.~\ref{ssec:results}) show that the alignments learned from the NMT training are sufficient and we do not lose translation performance.

\section{Experiments}
We test our vocabulary selection approach on two language pairs: German$\rightarrow$English and Italian$\rightarrow$English.
The alignments from which we extract the candidate lists are learned either during the full training (from scratch) or only during the final epoch (continue training).

\subsection{Setup}
For the German$\rightarrow$English translation task, we train an NMT system based on the WMT 2016 training data~\cite{bojar2016findings} (3.9 parallel sentences) and
use newstest-2014 and newstest-2015 as our dev/ test sets.
For the Italian$\rightarrow$English translation task, we train our system on a large data set consisting of 20 million parallel sentences. The sentences come from varied resources such as Europarl,
news-commentary, Wikipedia, openSubtitles among others. As test set, we use Newstest-2009.

In all our experiments, we use our in-house attention-based NMT implementation which is similar to~\cite{bahdanau+:2014}.
We use sub-word units extracted by byte pair encoding~\cite{sennrich2015neural} instead of words,
which shrinks the vocabulary to 40k sub-word symbols for both source and target.
For comparison, we also run a word model only for German$\rightarrow$English. We limit our word vocabularies to
be the top 80K most frequent words for both source and target.
Words not in these vocabularies are mapped to a single \textit{unknown} token. The oov-rate for the 80K word-based model on the dev and test sets is about $4.5\%$ and $4.3\%$ respectively.
During translation, we use the alignments (learned from the attention mechanism) to replace the unknown tokens either with
potential targets (obtained from an Model-1 trained on the parallel data) or with the source word itself (if no target was found).
For all our experiments, we use an embedding dimension of 620 and fix the recurrent GRU layers to be of 1000 cells each. For
the training procedure, we use uAdam~\cite{kingma2014adam} to update model parameters with a mini-batch size of 80.
The training data is shuffled after each epoch.

For evaluation, we compare the \bleu and \ter scores for the baseline decoding and the vocabulary selection decoding. For the baseline decoding, we use full search without using the candidate vocabulary from the learned alignments. We further compare the candidate lists of different sizes, where we limit the maximum number of target words per source word to be $20$, $50$, $100$ or $200$. For the continued training setup, we train models to convergence without learning alignments and then train one epoch to learn the alignments.
For decoding speed comparison, we report relative speedup gains with respect to the baseline, by averaging across 10 runs.% We also show the average number of candidate vocabulary generated from the alignments for different candidate list sizes.

\subsection{German$\to$English}
\label{ssec:results}

The results for the German$\rightarrow$English translation task are shown in Table~\ref{tab:deen-results}. Applying vocabulary selection during decoding speeds up the decoding by up to $7$x compared to the baseline decoding without any vocabulary selection. In our experiments across both languages, candidate list of size $100$ best target candidates per word, seems to be a good trade-off setting between speedup gain (over $3$x) without loosing any performance. The average numbers of candidates (per source word) in this case is just a tiny fraction ($\approx 0.1\%$) of the full target vocabulary used in the baseline setting.

One of the interesting trends, we note is that continue training turns out to be extremely competitive to learning alignments throughout NMT training over several epochs. It should be stressed again that we only ran the continue training setup for one epoch to learn alignments. This suggests that the attention weights are very stable once the model is reasonably trained. While the word based models are slightly worse than the BPE models, we observe the same trends that we noted earlier in terms of speedup and number of candidates.

For both BPE and word-based models, we also compare our vocabulary selection with one generated from traditional IBM models typically used. We employ vocabulary selection from alignments generated by \textit{fastalign}~\cite{dyer+:2013}, which is reparameterized model based on IBM Model-2. We limit the number of top-$n$ target words to be 100 to match our chosen setting. As can be seen in the table (last rows in BPE and Words blocks), the \bleu and \ter scores are similar to the numbers using our approach for the same top-$n$ setting. However, the average number of candidates from the fastalign is significantly larger: about $10\%$ for BPE and $30\%$ for words. We plan to compare the target candidates from both approaches in future and we also hope that, it would give us some insights to improve our approach.

\begin{table*}[t!]
    \centering
    \begin{tabular}{|l|c|c|c|c|c|c|}
        \hline
\bf{Model/ Vocabulary} & \bf{Alignments} & \bf{Cand list} & \bf{Speedup} & \bf{Avg cands} & \multicolumn{2}{c|}{\bf{Newstest-2015}} \\ \cline{6-7}
\bf{Size} & \bf{Learning} & \bf{Size} & \bf{Gain} & \bf{per word} & \BLEU & \TER \\ \hline \hline
\multirow{8}{*}{BPE ($\approx 40K$)} & No alignments & - & - & 34,494 & \rdm{27.87} & \rdm{52.88}\\ \cline{2-7} 
& \multirow{4}{*}{From scratch} & 20 & \rdm{3.31}x & \rdm{11.3167} & \rdm{27.13} & \rdm{53.67}\\ \cline{3-7}
& & 50 & \rdm{3.191}x & \rdm{24.4678} & \rdm{27.47} & \rdm{53.33}\\ \cline{3-7}
& & 100 & \rdm{3.096}x & \rdm{43.0317} & \bf{\rdm{27.77}} & \bf{\rdm{53.00}} \\ \cline{3-7}
& & 200 & \rdm{2.945}x & \rdm{74.6267} & \rdm{27.81} & \rdm{52.96}\\ \cline{2-7}
& \multirow{4}{*}{Continue training} & 20 & \rdm{3.239}x & \rdm{10.2652} & \rdm{27.08} & \rdm{53.82}\\ \cline{3-7}
& & 50 & \rdm{3.167}x & \rdm{22.3045} & \rdm{27.48} & \rdm{53.27}\\ \cline{3-7}
& & 100 & \rdm{3.088}x & \rdm{38.5956} & \bf{\rdm{27.71}} & \bf{\rdm{53.10}} \\ \cline{3-7}
& & 200 & \rdm{2.969}x & \rdm{66.3965} & \rdm{27.79} & \rdm{52.96} \\ \cline{2-7}
& Fast-align & 100 & \rdm{3.072}x & \rdm{42.282559} & \rdm{27.92} & \rdm{52.90} \\ \hline \hline
\multirow{8}{*}{Words (80K)} & No alignments & - & - & 80,000 & \rdm{26.24} & \rdm{54.28}\\ \cline{2-7}
& \multirow{4}{*}{From scratch} & 20 & \rdm{7.275}x & \rdm{7.3469} & \rdm{26.14} & \rdm{54.15}\\ \cline{3-7}
& & 50 & \rdm{7.156}x & \rdm{17.4309} & \rdm{26.31} & \rdm{54.03}\\ \cline{3-7}
& & 100 & \rdm{6.93}x & \rdm{31.8184} & \bf{\rdm{26.53}} & \bf{\rdm{53.90}} \\ \cline{3-7}
& & 200 & \rdm{6.704}x & \rdm{56.8326} & \rdm{26.54} & \rdm{53.92} \\ \cline{2-7}
& \multirow{4}{*}{Continue training} & 20 & \rdm{7.302}x & \rdm{6.9129} & \rdm{26.15} & \rdm{54.16} \\ \cline{3-7}
& & 50 & \rdm{7.196}x & \rdm{16.0271} & \rdm{26.36} & \rdm{53.99} \\ \cline{3-7}
& & 100 & \rdm{7.023}x & \rdm{28.3114} & \bf{\rdm{26.55}} & \bf{\rdm{53.93}} \\ \cline{3-7}
& & 200 & \rdm{6.811}x & \rdm{49.4804} & \rdm{26.56} & \rdm{53.94} \\ \cline{2-7}
& Fast-align & 100 & \rdm{6.943}x & \rdm{36.6832} & \rdm{26.66} & \rdm{53.95} \\ \hline
    \end{tabular}
    \caption{Vocabulary Selection for decoding: German$\rightarrow$English test set (newstest-2015) \bleu and \ter, with $\alpha_{thr} = 0.1$. Speedup gain is the relative gain compared to the baseline decoding without candidate list (computed on CPU). Average candidates refers to the average number of unique target vocabulary items per source word.}
    \label{tab:deen-results}
\end{table*}

\begin{figure*}
    \centering
    \begin{subfigure}[b]{0.49\textwidth}
        \includegraphics[width=\textwidth]{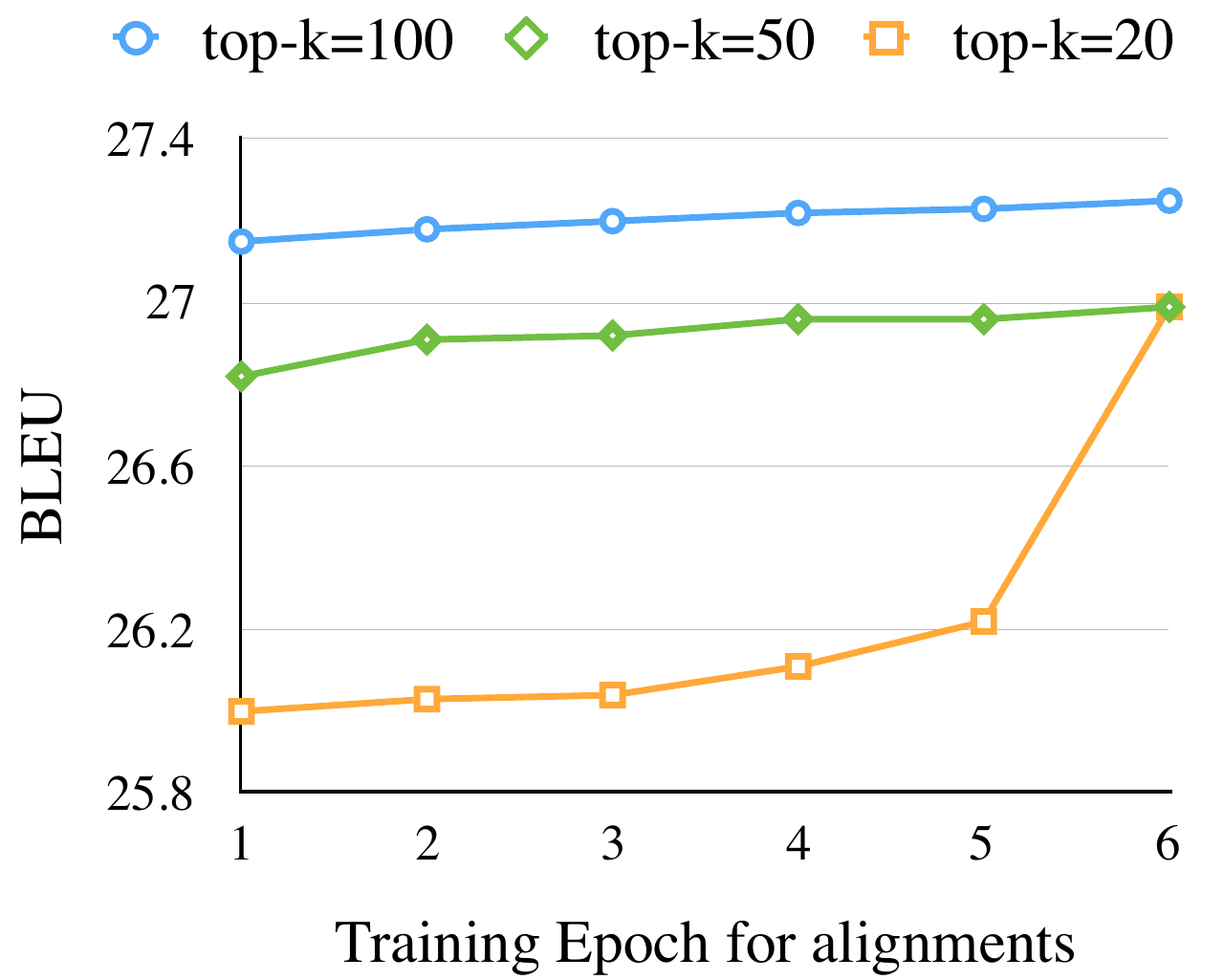}
        \caption{\bleu across different epochs}
        \label{fig:bleu_across_epochs}
    \end{subfigure}
    ~ %add desired spacing between images, e. g. ~, \quad, \qquad, \hfill etc. 
    \begin{subfigure}[b]{0.49\textwidth}
        \includegraphics[width=\textwidth]{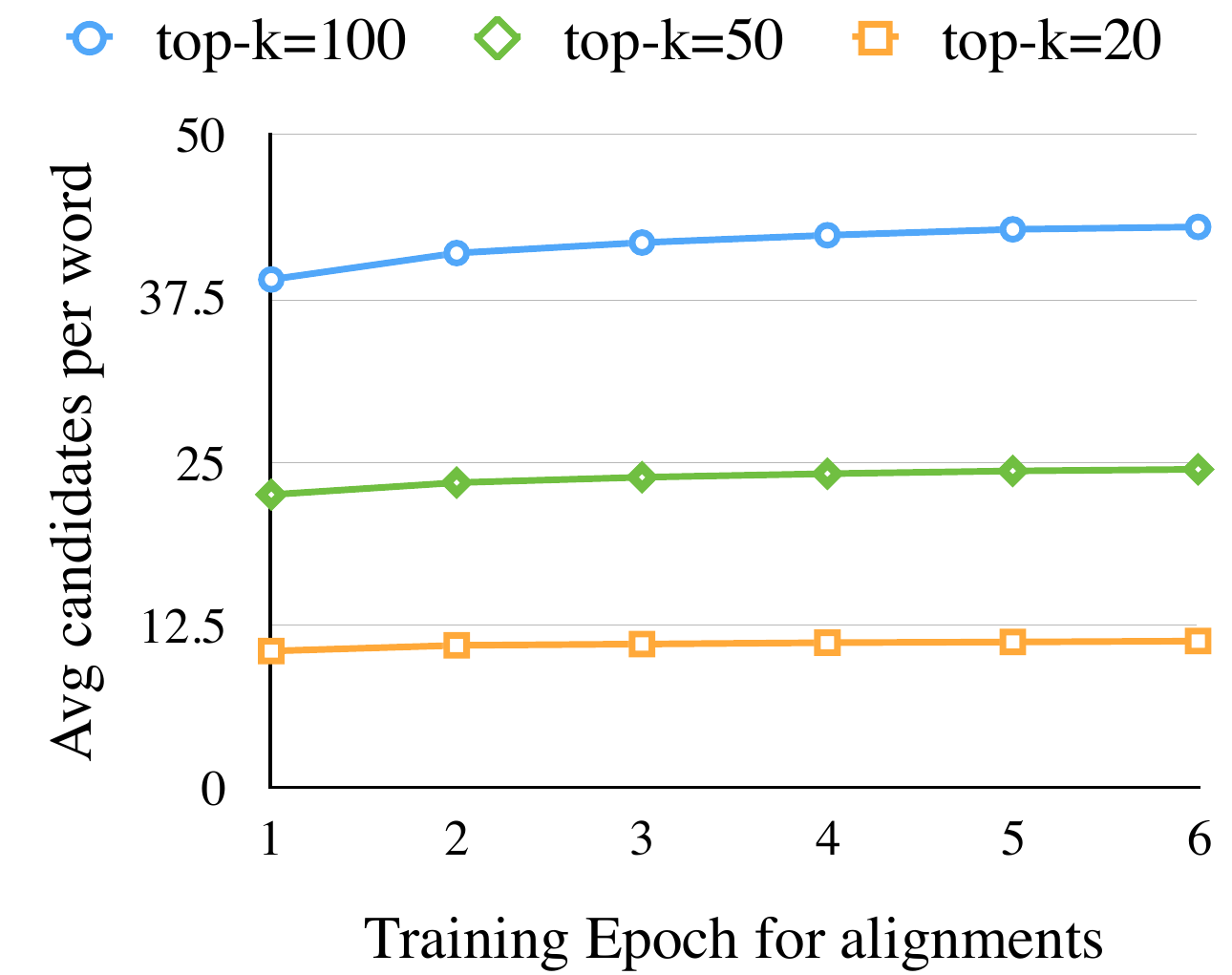}
        \caption{Avg. \# of cands across different epochs}
        \label{fig:cands_across_epochs}
    \end{subfigure}
    %~ %add desired spacing between images, e. g. ~, \quad, \qquad, \hfill etc. 
    \caption{Effect of alignments learned for different number of epochs in German$\rightarrow$English setting (dev set: newstest-2014). The decoding is performed by fixing the NMT model and only changing the alignments, which are obtained from the training of corresponding epoch in the x-axis. Figure~\ref{fig:bleu_across_epochs} on left correspond to \bleu and ~\ref{fig:cands_across_epochs} on right to average, unique target candidates (plots in broken lines). Easier to read in color.}\label{fig:aligns_at_epochs}
\end{figure*}

We show the effect of learning alignments for different number of epochs for German$\rightarrow$English translation setting in Figure~\ref{fig:aligns_at_epochs} for our devset (newstest-2014). In these experiments, we fix the NMT model and only change the source-target word distribution for vocabulary selection during decoding. The left plot shows the effect on \bleu scores while the one on the right-side plots average number of candidates per source word. The word distribution from early epochs seems to negatively affect the smaller candidate lists (top-$20$ in figure), where the \bleu increases by one point. For larger candidate sizes, the effect is only marginal in \bleu ($\approx 0.2$). As for the candidates, we see a flat curve, when the candidate list size is small (say $20$ or even $50$) and it starts to have some variance for 100 or more candidates.

\begin{table*}[t!]
    \centering
    \begin{tabular}{|l|c|c|c|c|c|c|}
        \hline
\bf{$\alpha$ thresholds} & \bf{Avg cands} & \multicolumn{2}{c|}{\bf{Newstest-2014}} & \multicolumn{2}{c|}{\bf{Newstest-2015}} & \bf{Density/} \\ \cline{3-6}
 & \bf{per word} & \BLEU & \TER & \BLEU & \TER & \bf{Size (MB)} \\ \hline \hline
$\alpha_{thr} = 0.05$ & \rdmd{43.7283} & \rdmd{27.25} & \rdmd{54.12} & \rdmd{27.76} & \rdmd{52.98} & \rdmt{3.0539}/ \rdm{172.618852} \\
$\alpha_{thr} = 0.1$ & \rdmd{43.6759} & \bf{\rdmd{27.25}} & \bf{\rdmd{54.12}} & \bf{\rdmd{27.77}} & \bf{\rdmd{53.00}} & \rdmt{1.9278}/ \rdm{108.9676} \\
$\alpha_{thr} = 0.2$ & \rdmd{40.2224} & \rdmd{27.27} & \rdmd{54.06} & \rdmd{27.7} & \rdmd{53.09} & \rdmt{1.011399}/ \rdm{57.16758} \\
$\alpha_{thr} = 0.25$ & \rdmd{39.0165} & \rdmd{27.23} & \rdmd{54.16} & \rdmd{27.76} & \rdmd{53.05} & \rdmt{0.80936}/ \rdm{45.748024} \\ \hline
    \end{tabular}
    \caption{German$\rightarrow$English: Effect of different $\alpha_{thr}$ for accumulating alignments (with candidate list size = 100). Density refers to the percent of non-zero entries among the full size of the source-target vocabulary matrix and size indicates the raw size for the non-zero entries (in MB).}
    \label{tab:deen-alpha-threshold}
\end{table*}

Table~\ref{tab:deen-alpha-threshold} shows the effect of different $\alpha_{thr}$ thresholds for accumulating the alignments. The $\alpha_{thr}$ could be used to strike a balance between desired coverage in source-target word distribution and avoiding spurious source-target links. We observe that the thresholds up to $0.25$ result in similar performance levels (shown for both dev and test sets), with smaller candidate vocabulary size as the threshold is increasing. We also noticed at larger thresholds, the accumulated count matrices lacked variety in the source words distribution, leading to poor coverage. This is to be expected because, for such large thresholds, the alignments will be accumulated only when attention exhibits a peaked distribution it that it is strongly confident about some particular source-target link. We believe $0.1$ or $0.2$ would be practical $\alpha_{thr}$ values for most data sets/ language pairs.

\subsection{Italian$\to$English}
\label{ssec:results-iten}
Empirical results for the Italian$\to$English translation task are shown in Table~\ref{tab:iten-results}.
We can speed up the decoding speed by a factor of 3.6x to 3.9x by using a candidate list coming from the attention of our NMT model. The sweet spot candidate size is 100. We can speed up the decoding by a factor of 3.7 while losing only 0.1 point in \BLEU.
The continue training (in which we only learn the candidate list in the final epoch), works as good as the full trained candidate list. The average candidate per words are even smaller compared to the full trained candidate list which makes the decoding even
a little bit faster.

\begin{table*}[t!]
    \centering
    \begin{tabular}{|c|c|c|c|c|c|}
        \hline
\bf{Alignments} & \bf{Cand list} & \bf{Speedup} & \bf{Avg cands} & \multicolumn{2}{c|}{\bf{Newstest-2009}} \\ \cline{5-6}
\bf{Learning} & \bf{Size} & \bf{Gain} & \bf{per word} & \BLEU & \TER \\ \hline \hline
No cand list & - & - & 33 497 & \rdm{29.73} & \rdm{52.68} \\ \cline{1-6} %14.9
\multirow{4}{*}{From scratch} & 20  & 3.9x & 10.6 & \rdm{28.55} & \rdm{52.64} \\ \cline{2-6} %57.1
& 50  & 3.8x & 23.6 & \rdm{29.21} & \rdm{53.47} \\ \cline{2-6} %56.3
& 100 & 3.7x & 42.9 & \bf{\rdm{29.55}} & \bf{\rdm{53.14}} \\ \cline{2-6} %55.3
& 200 & 3.6x & 76.8 & \rdm{29.62} & \rdm{52.98} \\ \cline{1-6} %53.6
\multirow{4}{*}{Continue training} & 20 & 3.9x & 10.1 & \rdm{28.62} & \rdm{52.61} \\ \cline{2-6} %56.8
& 50 & 3.8x & 22.4 & \rdm{29.21} & \rdm{53.48} \\ \cline{2-6} %56.1
& 100 & 3.7x & 40.9 & \bf{\rdm{29.51}} & \bf{\rdm{53.08}} \\ \cline{2-6} %54.9
& 200 & 3.6x & 72.7 & \rdm{29.65} & \rdm{52.97} \\ \hline %53.1
    \end{tabular}
    \caption{Results for Italian$\to$English translation task, with $\alpha_{thr} = 0.1$.}

    \label{tab:iten-results}
\end{table*}

\subsection{Dynamic Vocabulary Selection during Training}
\label{ssec:dynamic_voc_manip}
As we accumulate the attention weights into a sparse alignment matrix, we could also exploit this to dynamically select the target vocabulary during training. This would be exactly same as the large vocabulary NMT; but unlike other approaches we would not be relying on external resource/ tools such as Model-1 alignments, phrase tables etc.

We now explain the recipe for doing this. We first normalize the sparse matrix and obtain the top-$n$ target tokens for each source word as explained in Section~\ref{ssec:decoding_voc_manip}.\footnote{In order to avoid stale probabilities, we normalize/ trim the alignments at the beginning of each epoch for dynamic vocabulary selection.}

We begin the NMT training without any vocabulary selection and train with entire target vocabulary during the initial stages. We switch to vocabulary selection mode, once the alignment matrix is seeded with initial alignments from at least one full sweep over data. Given a mini batch of source sentences $\mathcal{X}$, we identify for each source word, top-$n$ target words and use the set of all unique target words $\mathscr{Y}_c$ as candidate vocabulary for that batch.

\begin{equation}
\mathscr{Y}_c = \{ f_a(x_i) \} \; \forall x_i \in \mathcal{X}
\end{equation}

The dynamic vocabulary can then be used as the target vocabulary to train the present batch. Dynamic selection for each mini batch during training could add to the computational cost and potentially slow it down. One simple solution would be to just do an offline vocabulary selection based on the alignments at the beginning of each epoch. We leave this for future experimentation.

\section{Related Works}
Vocabulary selection has been studied widely in the context of NMT decoding~\cite{luong+:2015a,mi+:2016a,lhostis+:2016}. All these works are inspired by the early work by~\namecite{jean+:2015} and use some kind of external strategy (based on word alignments or phrase tables or co-occurrence counts etc.) in order to do vocabulary selection. In contrast, we use the alignments that are learned from the attention weights in the early training, for selecting target vocabulary. The other difference is that the selected vocabulary remains stale throughout the training under these earlier approaches. However in this work, the alignments learned in the previous epoch could be used to select target vocabulary for next epoch.

Hierarchical softmax \cite{morin2005hierarchical,mnih2009scalable} is well-known way to reduce softmax over a large number of target words. It uses a hierarchical
binary tree representation of the output layer with all words as its leaves.  It allows
exponentially faster computation of word probabilities and their gradients, but the predictive
performance of the resulting model is heavily dependent on the tree used,  which is often constructed heuristically.
Moreover, by relaxing the constraint of a binary structure,~\namecite{le2011structured} and~\namecite{baltescu2014pragmatic}
introduce a structured output layer with an arbitrary tree structure constructed
from word clustering. All these methods speed up both the model training and evaluation considerably but are heavily depend on the quality of the word cluster. NMT experiments with hierarchical softmax showed improvement for smaller datasets with about 2m sentence pairs~\cite{baltescu2014pragmatic}.

\section{Summary}
We presented a simple approach for directly learning the source-target alignments from the attention layer in Neural Machine Translation. We showed that the alignments could be used for vocabulary selection in decoding, without requiring any external resources such as aligners, phrase-tables etc. We recommend setting $\alpha_{thr} = 0.1$ and top-$n$ candidates to be $100$ for good performance and faster decoding for most language pairs/ datasets. Our experiments showed decoding speedup of up to a factor of $7$ for different settings. We also showed how this could be used for dynamic vocabulary selection during training.

%\balance
%\newpage
%\bibliography{acl2017}
\bibliography{candList}
\bibliographystyle{acl_natbib}

\appendix

\end{document}